\pdfoutput=1
\documentclass[11pt]{article}

\usepackage[]{acl}

\usepackage{times}
\usepackage{latexsym}
\usepackage{microtype}
\usepackage{graphicx}
\usepackage{subfigure}
\usepackage{booktabs} % for professional tables
\usepackage{amsmath} % for equations
\usepackage{tabularx}
\usepackage{multirow}
\usepackage{float}
\usepackage{adjustbox}

\usepackage[T1]{fontenc}
\usepackage[utf8]{inputenc}

\usepackage{microtype}

\title{Ticket-BERT: Labeling Incident Management Tickets with Language Models}

\author{Zhexiong Liu$^{1}$\thanks{\text{  } The work was done while interning at Microsoft.}\text{  }, Cris Benge$^2$, Siduo Jiang$^2$ \\
  $^1$Department of Computer Science, School of Computing \& Information \\
  University of Pittsburgh, Pittsburgh, PA 15260 \\ $^2$Microsoft Azure, Redmond, WA 98052  \\
  \texttt{zhexiong@cs.pitt.edu} \texttt{\{crbenge,stonejiang\}@microsoft.com} \\
  }

\begin{document}
\maketitle

\begin{abstract}
\label{sec:abstract}
% Software and hardware incident management has gained notable attention in enterprise-level products, where efficient ticket resolution services are critically needed due to large-scale ticket resolution demands. 
An essential aspect of prioritizing incident tickets for resolution is efficiently labeling tickets with fine-grained categories. 
% This facilitates the accurate triage of incidents to responsible teams and ultimately accelerates ticket resolution. 
However, ticket data is often complex and poses several unique challenges for modern machine learning methods: $\left(1\right)$ tickets are created and updated either by machines with pre-defined algorithms or by engineers with domain expertise that share different protocols,  $\left(2\right)$ tickets receive frequent revisions that update ticket status by modifying all or parts of ticket descriptions, and $\left(3\right)$ ticket labeling is time-sensitive and requires knowledge updates and new labels per the rapid software and hardware improvement lifecycle. To handle these issues, we introduce Ticket-BERT which trains a simple yet robust language model for labeling tickets using our proposed ticket datasets.
% (i.e., D-Human, D-Machine, D-Mixture)
Experiments demonstrate the superiority of Ticket-BERT over baselines and state-of-the-art text classifiers on Azure Cognitive Services. 
% We further evaluate Ticket-BERT using human input on a set of hard-to-identify tickets and achieve outstanding performance.
We further encapsulate Ticket-BERT with an active learning cycle and deploy it on the Microsoft IcM system, which enables the model to quickly finetune on newly-collected tickets with a few annotations.
\end{abstract}
\section{Introduction}
\label{sec:introduction}
The explosive growth of digitization in enterprise-level products emphasizes the need for highly-reliable and functional services with efficient incident maintenance (e.g., efficiently resolving system interruptions or outrages). These problematic incidents are typically governed by incident management systems \cite{gupta2008automating,gupta2009multi}, where hardware and software failures are documented as tickets using a series of formative descriptions and arrays of structured fields (e.g., date, titles, keywords, severity). The essential aspect of processing incident tickets lies in labeling issues with specific tags that precisely exhibit incident categories \cite{herzig2013s,zhou2016combining,revina2020ticket}; however, this is often complicated by highly variable content, which makes it challenging to transfer knowledge among diverse environments. For example, the tickets could be automatically created by machines that monitor fatal errors or initiated by engineers. The mixture of human and machine-generated tickets has highly variable,  issue-specific vocabularies. Moreover, tickets may receive frequent revisions on the status and descriptions based on newly updated information (e.g., ticket acknowledgment, mitigation, remediation). New knowledge and labels may also be needed when software and hardware upgrade to new versions or when new issues occur. Therefore, developing a domain-adaptive ticket labeling system is necessary for achieving acceptable performance.

Ticket classification has been widely explored in machine learning tasks, such as ticket assignment \cite{mukunthan2019multilevel}, prioritization \cite{kallis2019ticket,revina2020ticket}, resolution \cite{ali2022multiapproach}, and duplicate identification \cite{gu2011analysis}. These methods generally analyze an incident ticket solely by using its cleaned content (e.g., title, description, structured data) rather than focusing on the raw process of ticket creation, updates, and resolution that convey rich and valuable ticket information. Typically, tickets undergo multiple updates throughout their lifecycle; details are added or corrected by potentially various authors over time. However, little work has been done leveraging this update-related information to address the task of ticket labeling. Moreover, the existing machine learning approaches \cite{kallis2019ticket, maksai2014hierarchical, son2014automating, han2018vertical} train ticket classification models with small-size domain-specific datasets that are not transferable and generalizable to large-scale real-world problems. To bridge these gaps, we develop new datasets using 76K raw tickets and ten fine-grained issue-specific labels. Specifically, each ticket receives dozens of updates in its lifecycle, and we utilize a few of those to create D-Machine, D-Human, and D-Mixture datasets containing ticket titles, descriptions, and summaries generated by machines, engineers, and both, respectively. Afterward, we develop language models to identify issue-related ticket labels on these datasets. Note that our datasets are developed based on the first five ticket updates; thus, our trained model can label tickets as soon as they are created. 
% This significantly helps label new tickets in their early stage and ultimately accelerates ticket resolution.

Adapting machine learning models to text classification tasks has gained unprecedented interest in recent research \cite{kowsari2019text, kadhim2019survey}. Multiple text representation techniques and classification methods have been proposed \cite{kim1992fast,sidhu2001fast,hadi2018integrating, yao2019clinical}. 
% Typically, text classification involves three stages: $\left(1\right)$ feature extraction that utilizes salient information in raw text to present text semantics, such as TF-IDF and Bag-of-Words (BoW) features, $\left(2\right)$ data down-sampling that aims to reduce the dimensions of text vector representations or leverage fewer samples to train lightweight classifiers that can be efficiently deployed in low memory systems, and $\left(3\right)$ supervised or unsupervised model training that either leverages many class annotations to conduct text classification or employs cluster-based methods to group similar text \cite{agarwal2014text}. 
Generally, these methods often perform poorly when the text data (e.g., raw tickets) are not clean or have large variability. For example, the ticket descriptions regarding software incidences would be significantly different from those that address hardware incidences in terms of text semantics and syntactic structures. To address this variability, end-to-end and domain-adaptive methods are needed. In a previous study, transformer-based language models have demonstrated robustness in learning context-aware text information from diverse environments \cite{9373074,khan2021transformers}, but their applications in ticket labeling tasks remain relatively unexplored. A scarcity of labeled data for issue-specific incidences in real-world ticketing systems has limited the ability to develop effective deep learning models for classification. 
% We overcome this challenge by introducing large-scale datasets and the first high-performance Ticket-BERT.

Against this backdrop, we study various ticket labeling classification methods and demonstrate our language models' superiority over a set of strong baselines on all the proposed ticket datasets. Our contribution is fourfold: $\left(1\right)$ We develop the first large-scale real-word ticket dataset that contains ticket titles, summaries, and multiple updates of ticket descriptions, $\left(2\right)$ We develop highly-effective language models that gain state-of-the-art performance on all the proposed datasets over multiple strong baselines, including Microsoft Cognitive Service, $\left(3\right)$ We develop a novel text sampling strategy that leverages auxiliary text to greatly boost ticket labeling performance on the difficult D-Human dataset, and $\left(4\right)$ we proposed an active-learning pipeline in the real incident management systems that iteratively learns models from new data. This ensures domain adaptation for the Ticket-BERT when handling diverse incidents.
\section{Data Collection}
\label{sec:dataset}
\subsection{Data Source}
\begin{figure*}[t]
    \centering
    \includegraphics[width=0.97\textwidth]{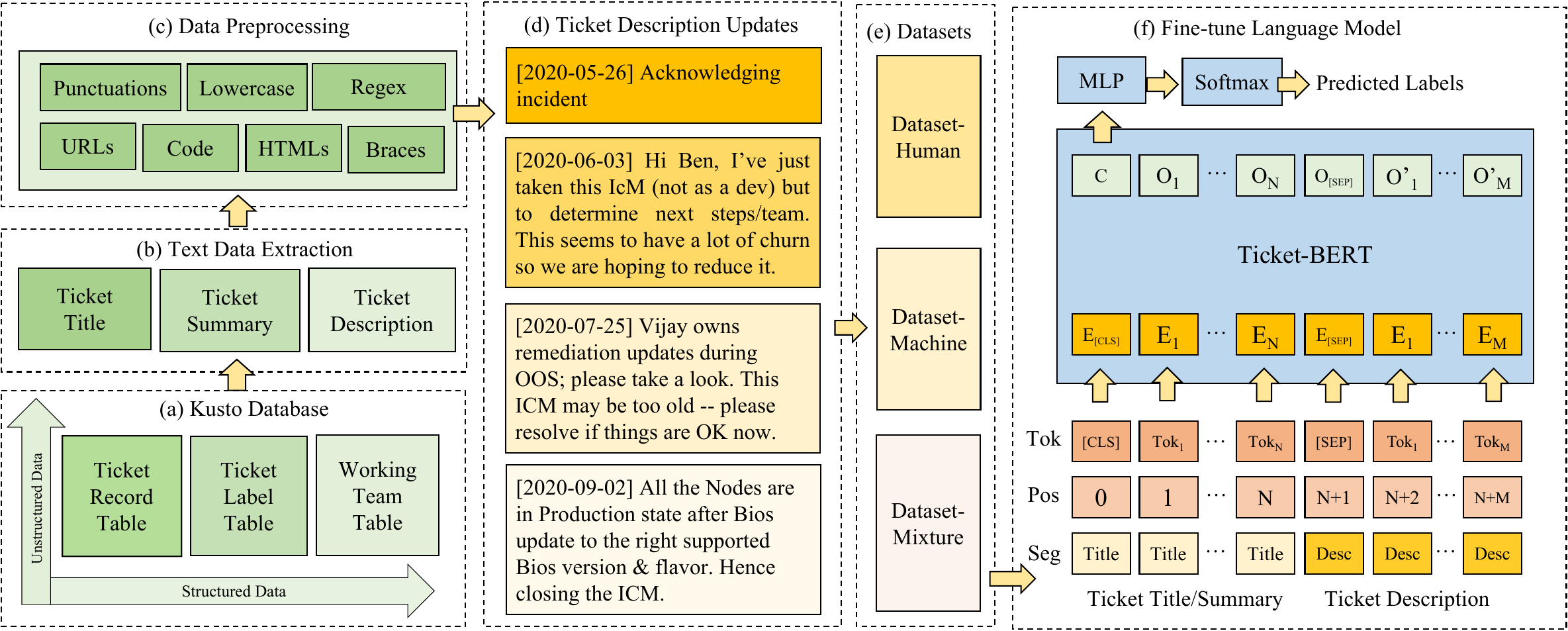}
    \vspace{-.05in}
    \caption{Framework: a) Kusto database that stores structured and unstructured ticket data. (b) Pulled 76K raw tickets from Kusto, including ticket titles, summaries, and descriptions. (c) The tickets are preprocessed using predefined rules. The clean tickets in (d) show four out of dozens of ticket updates that reflect the new ticket status. (e) build three datasets using ticket descriptions: D-Human where ticket descriptions are entered by humans, D-Machine where ticket descriptions are generated by machines, D-Mixture where it is a mixture of D-Human and D-Machine (Sec. \ref{sec:dataset}). (f) Fine-tune a Ticket-BERT using ticket title/summary as an auxiliary prompt prefix concatenated on ticket descriptions (Sec. \ref{sec:methods}).}
    \vspace{-.01in}
    \label{fig:framework}
\end{figure*}

Presently, we are not aware of a labeled large-scale dataset for incident management. As a result, we curated the first ticket dataset using 76K raw tickets pulled from Microsoft Kusto\footnote{\url{https://docs.microsoft.com/en-us/azure/data-explorer/kusto/query/}} between June 2020 and June 2022. The raw data contains 5 fields, including Title, Description, Summary, Edited by, Timestamp. To preprocess the data, we cleaned URLs, Code, HTML Tags, Tables, and other metadata, in order to parse out plain text in the ticket in titles, descriptions, and summaries. Given that each ticket $T$ receives multiple ($m$) modifications in their lifecycles, as shown in Figure \ref{fig:framework} (d), either from machines or from engineer, we define the ticket update $T_i$ as the $i$-th record of modification on the ticket description.  For example, Table \ref{tab:ticket_timeline} in Appendix \ref{apdx:some_section} shows the parts of ticket updates ranging from May 2020 to September 2020 which suggests it is a complex incident that costs months to resolve.  Typically, a human-entered/updated $T_i$ has high variability and often addresses diverse incident issues with specific details. Machine-generated/updated $T_i$ focuses on a smaller number of incident failures that result from automation. Table \ref{tab:ticket_examples} in Appendix \ref{apdx:some_section} shows the human-entered $T_i$ are close to natural language full of transition and reasoning, while machine-generated $T_i$ sharing similar patterns but have low variability. 

\subsection{Dataset Construction}
Note that each ticket has multiple description updates $T_i$ where $i=1,2,\dots,m$, and all the updates describe the same incident issues in different timelines. For simplicity, we draw one out of $m$ updates as a representative description of each ticket to build our datasets. Specifically, we drop short ticket descriptions that have no salient semantic information about incident issues, such as \textit{``incident acknowledged"}. Here the short ticket descriptions are defined as text with less than $n$ characters. To investigate the best value of $n$, we show the frequency distribution of $T_i$ with respect to its length ($n$ characters) in Table \ref{tab:ticket_length}. Interpreting this table, 99.7\% of data is drawn from the first update $T_1$ (0.1\% from the second update $T_2$) of the ticket descriptions if we draw the first long text with more than $n=10$ characters. In dataset construction, we develop D-Human dataset that only draws ticket descriptions entered by humans, D-Machine that only draws ticket descriptions generated by machines, and D-Mixture that is a hybrid version of D-Human and D-Machine. Specifically, we set $n=50$ which ensures the collected ticket descriptions are mostly from the first 3 descriptions ($T_1$, $T_2$, $T_3$). This way, the models trained on these datasets are able to label tickets as soon as the tickets are created, which significantly helps identify new ticket labels in their early stage (e.g., first 5 updates) and ultimately help accelerate ticket resolution.

\begin{table}[]\small
\centering
\begin{adjustbox}{width=1\linewidth}
\begin{tabular}{l|cccccc}
\hline
\multirow{2}{*}{\begin{tabular}[c]{@{}l@{}} Text \\ Length ($n$)\end{tabular}} & \multicolumn{6}{c}{Ticket description updates ($T_i$)}                 \\ \cline{2-7} 
  & $T_1$    & $T_2$   & $T_3$   & $T_4$   & $T_5$   & others \\ \hline
10 (char)    & 99.7\% & 0.1\%  & 0.0\% & 0.0\% & 0.0\% & 0.2\%  \\
20  (char)  & 30.7\% & 68.8\% & 0.1\% & 0.0\% & 0.0\% & 0.2\%  \\
50 (char)    & 29.3\% & 65.6\% & 3.1\% & 0.7\% & 0.2\% & 1.0\%  \\
100 (char)    & 27.1\% & 64.9\% & 3.1\% & 1.5\% & 0.6\% & 2.8\%  \\
200  (char)  & 25.7\% & 64.1\% & 2.3\% & 1.4\% & 0.9\% & 5.6\%  \\ \hline
\end{tabular}
\end{adjustbox}
% \vspace{-.1in}
\caption{The percentage frequency distribution of drawn examples that are from $T_i$-th update of the ticket description with longer than $n$-character.}
\vspace{-.2in}
\label{tab:ticket_length}
\end{table}

\subsection{Data Distribution}
In order to specifically identify incident issues, we develop 10 fine-grained ticket labels (Table \ref{tab:ticket_labels} in Appendix \ref{apdx:some_section}). Figure \ref{fig:label_distribution} shows label distribution in D-Human dataset, where \textit{CloudNet FPGA Alerts} issues are the most. These labels were developed based on feedback from subject matter experts within the Azure Hardware space. Primarily, the focus was on identifying systematic issues, and then bucketing these issues based on the previous root cause, or potential path to resolution. The ticket labels shown are the best estimates for potential problems underlying our IcM tickets. In this dataset, we only have 10 labels total for all tickets. Statistically, D-Human contains 20,142 tickets for training, 2,239 for validation, and 5,596 for testing. D-Machine contains 39,391 tickets for training, 4,377 for validation, and 10,942 for testing. D-Mixture contains 40,611 tickets for training, 4,513 for validation, and 11,282 for testing.

\begin{figure}[t]
    \centering
    \includegraphics[width=1\linewidth]{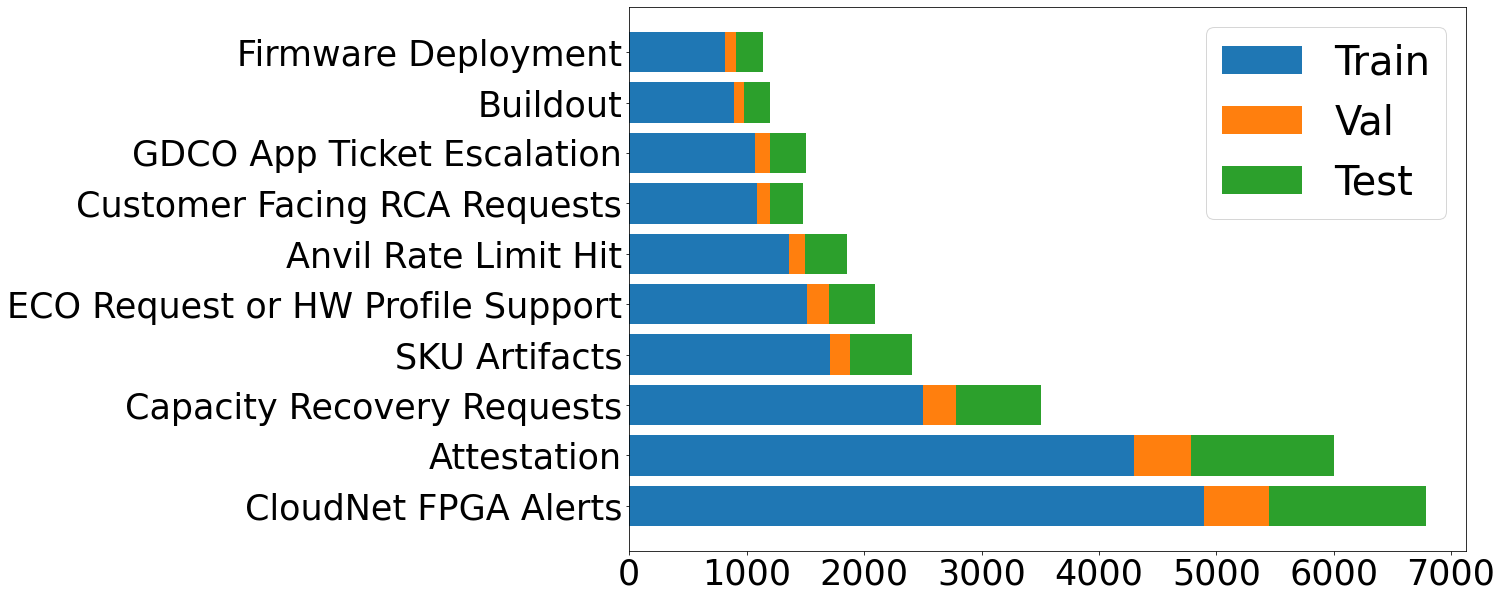}
    % \vspace{-.2in}
    \caption{Incident label distribution (numbers) in D-human dataset.}
    \vspace{-.15in}
    \label{fig:label_distribution}
\end{figure}

\section{Methods}
\label{sec:methods}

In this section, we introduce the ticket labeling framework shown in Figure \ref{fig:framework}, where we develop new ticket datasets and fine-tune Ticket-BERT on proposed dataset and achieve state-of-the-art performance. In particular, We fine-tune the BERT-base \cite{devlin2018bert} models with additional multi-layer perception and a softmax layer for multiclass classification, as shown in Figure \ref{fig:framework} (f), on the proposed datasets for multi-class classification. For baselines, we leverage TF-IDF and BoW features to train Naive Bayes and Logistic Regression models, and other competitive baseline classifiers on the Language Studio at Azure Cognitive Science\footnote{\url{https://language.cognitive.azure.com}}.

\subsection{Baseline Models}

\vspace{-.05in}
\textbf{NB-BoW}: Naive Bayes used in \cite{revina2020ticket} with Bag-of-Word (BoW) features. We construct a vocabulary containing all unique words found in the training ticket text and use a binary encoder to represent the word presence or absence. 
% In particular, a word is assigned one if in the vocabulary; otherwise, zero. We use Naive Bayes, which is demonstrated as an effective IT ticket classifier in \cite{revina2020ticket}, as our ticket labeling baseline.   

% \vspace{-.05in}
\textbf{NB-TF-IDF}: Naive Bayes with TF-IDF features. The Term Frequency-Inverse Document Frequency (TF-IDF) is a statistical measure that evaluates how relevant a word is to a document in a collection of documents. 
% The first metric is word frequency in a document and the second one is the inverse document frequency of the word. 
% In the ticket labeling settings, the document is a ticket description. The TF-IDF features are fed to a Naive Bayes model.  

% \vspace{-.05in}
\textbf{LG-BoW}: Logistic Regression with BoW features. Logistic Regression is often used in ticket classification \cite{revina2020ticket, 9836250} due to its light and explainable model structure. 
% 8768734
% We use BoW features as the model inputs. 
Please note that we use one-against-the-rest strategy 
% \cite{li2018multi}
to perform multi-class classification  on the logistic regression model.

% \vspace{-.05in}
\textbf{LG-TF-IDF}: Logistic Regression with TF-IDF features. For a fair comparison, we use the same TF-IDF features in NB-TF-IDF as another baseline.

% \vspace{-.05in}
\textbf{LS-Model}: Text classifiers on Microsoft Language Studio. To the best of our knowledge, the LS-Model is an advanced neural-network model, which serves as a competitive baseline. Please note that the reason why we benchmark on LS-Model rather than RNN/LSTM-based models is that $\left(1\right)$ the ticket descriptions are often longer than normal sentences (e.g, more than 50 words), which would make the RNN/LSTM-based models suffering from the issues of gradient vanishing or exploding. $\left(2\right)$ LS-Model is an commercialized and state-of-the-art text classification model which serves as a strong baseline.

% \vspace{-.05in}
\subsection{Ticket-BERT}
We fine-tune BERT-base model for the ticket labeling tasks on the proposed datasets. Instead of training BERT-base from scratch, we fine-tune pretrained BERT-base model which takes much less time to learn and achieve state-of-the-art results with minimal task-specific adjustments. The BERT-base model structure is shown in Figure \ref{fig:framework} (f). The inputs are a sequence of ticket descriptions. In order to leverage auxiliary data, such as ticket title and summary, we also develop prompt-prefix strategy to format input text sequences for training baseline and fine-tuning Ticket-BERT models. In particular, we concatenate prompt-prefix (ticket titles/summaries) and ticket descriptions in three templates: (1) \textit{[CLS] $<$description$>$};
(2) \textit{[CLS] $<$title$>$ [SEP] $<$description$>$};
(3) [CLS] $<$title$>$ [SEP] $<$summary$>$ [SEP] $<$description$>$. The formatted texts are inputs for fine-tuning the BERT-base model. These texts are encoded with word-piece tokens, positional embedding, and penitential segment embedding. All the encoded vectors are concatenated to feed the BERT-base model. Afterward, we extract the \textit{[CLS]} token embedding from the last layer in the BERT-base, and feed it to a Multi-layer Perceptron (MLP) to perform multi-class classification. In the fine-tuning stage, we load the initial BERT-base weights and only fine-tune the MLP block.  

\begin{figure}[t]
    \centering    \includegraphics[width=0.83\linewidth]{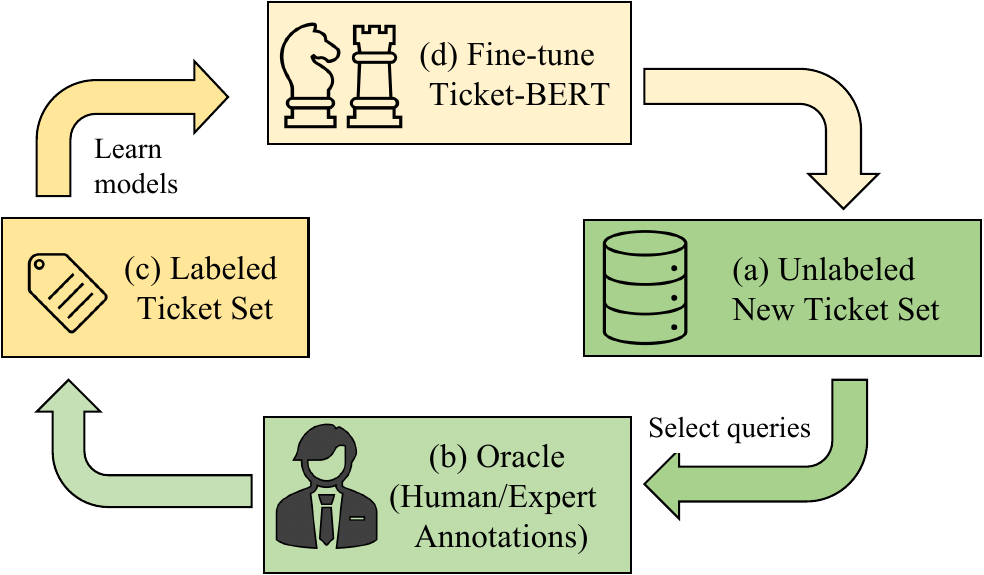}
    % \vspace{-.05in}
    \caption{The pool-based active learning cycle. (a) newly unlabeled tickets are collected; (b) a subset of queried unlabeled data requires human/expert annotations; (c) collected/extended ticket set with annotated labels; (d) fine-tune Ticket-BERT.}
    \vspace{-.1in}
    \label{fig:active_learning}
    % \vspace{-.1in}
\end{figure}

\subsection{Active Learning}

In the real-world ticket labeling task, a large number of unlabeled data can be collected every single day. For example, there are 7K new tickets created each day in Microsoft Kusto. These tickets could either cover new incidents that never existed before or address new issues (using new ticket labels) that was not focused on due to the rapid software and hardware improvement. Therefore an active learning \cite{lewis1995sequential} framework that could adapt previously learned models to the new data distribution is critically needed. To bridge this gap, we build a pool-based active learning cycle \cite{ settles2009active}, which assumes a small set of data $L$ is annotated but a large number of data $U$ has no labels. With this, we hope to achieve near-perfect accuracy as we continue to iterate with our customers. 

Suppose we learned an initial Ticket-BERT $M^{(i)}$ model using initially annotated data $L^{(i)}$, and new data $U^{(i)}$ is collected after fine-tuning $M^{(i)}$, where $i=0,1,2,..n$, we start a cycle of active learning in Figure \ref{fig:active_learning}: $\left(a\right)$ given new collected unlabeled data $U^{(i)}$, we use uncertainty-sampling strategy \cite{settles2009active} to query data from $U^{(i)}$. In particular, we use fine-tuned Ticket-BERT model $M^{(i)}$ to predict the possible labels with probabilities across instances in $U^{(i)}$, and query the instance $x^*$ whose best predicted labels has the least confident $x^*=\operatorname{argmin}_x P\left(y^*\mid x ; \theta\right)$
\noindent where $y^*=\operatorname{argmax}_y P(y\mid x ; \theta)$ is the best predicted label, $y$ are the candidate labels, $ x\in U^{(i)}$, $\theta$ is model weight. This strategy queries a set of the most uncertain or informative tickets $U^{(i)}_{q}$ from a large set of newly collected unlabeled data $U^{(i+1)}$. $\left(b\right)$ an oracle that involves human annotations on new queried data $L^{(i+1)}_{q}$. $\left(c\right)$ the annotated data augments existing labeled data to $L^{(i+1)}= L^{(i)} \cup L^{(i+1)}_{q}. $
\noindent $\left(d\right)$ $L^{(i+1)}$ is then used to further fine-tune Ticket-BERT and obtain $M^{(i+1)}$. Please note that the active learning cycle is launched on Microsoft IcM system and the process is ongoing. We will report new performance in the future work.
\section{Experiments and Analysis}
\label{sec:experiments}
\subsection{Implementation Details}
In the data preprocessing, we develop a set of rules to clean the text, such as remove URLs, Code, HTML Tags, Tables, Braces and other metadata. All the cleaned text are converted to lower cases and stopwords are removed. We conduct extensive experiments on the created datasets and achieve state-of-the-art performance. We implement the multi-class Naive Bayes (NB) and Logistic Regression (LR) with scikit-learn\footnote{\url{https://scikit-learn.org/stable/}} and Ticket-BERT with Pytorch\footnote{\url{https://pytorch.org/}}. We fine-tune the Ticket-BERT models on a 10-class classifier using cross-entropy losses on an Nvidia V100 GPU. The batch size is set to 16, and the learning rate is set to 0.0001, with 0.1 decay every 4 out of 12 epochs. In preprocessing, we trunk the text into 512 dimensions for the Ticket-BERT and set 1024 as the maximum dimensions for BoW and TF-IDF vectors. We report the best performance on macro Precision, Recall, F1, and AUC of ROC scores on the test set.

\subsection{Evaluation on the Dataset}
Table \ref{tab:results} shows the baselines and Ticket-BERT model performance. For the D-Human dataset, the Native Bayes models generally are worse than Logistic Regression models which suggests that discriminative modes are better than generative models in ticket labeling tasks. Moreover, the Logistic regression with BoW features is better than it with TF-IDF features; however, the Naive Bayes with TF-IDF features is better than with BoW features. This suggests that discriminative models favor features that cover the most of text information, such as BoW counting word occurrences, while generative models work well on salient words rather than the ones with high occurrences. In particular, the Ticket-BERT shows competitive results on D-Human in terms of precision, recall, and F1 scores, demonstrating the superiority of pretrained language models on ticket labeling. Regarding D-Machine and D-Mixture datasets, the Ticket-BERT has relatively lower recalls than the strongest baselines LG-BoW but shows better scores in terms of precision and F1 scores. This makes sense because D-Machine and D-Mixture datasets contain machine-generated/updated tickets that have relatively different semantic structures compared to human languages, shown in Table \ref{tab:ticket_examples} in Appendices.

\subsection{Evaluation on Prompt-prefix}
The model performance on D-Machine and D-Mixture is generally higher than that of in D-Human in Table \ref{tab:results}. Our reasoning is that the Human entered/updated ticket descriptions have rich variation that may confuse the machine learning models and make the task difficult, compared to machine-generated/updated text. To overcome these issues, we proposed the prompt-prefix that leverages auxiliary data (e.g., ticket title and summary) to enrich the ticket context and make the text more issue-specific. Table \ref{tab:results_prefix} shows the model performance after concatenating prompt-prefix. We observe strong boosts for all the models on the D-Human dataset and moderate improvement on the D-Machine and D-Mixture datasets if we take title as auxiliary text. While concatenating both summary and title on the D-Human dataset, Ticket-BERT's scores decrease from the one fine-tuned with title-only auxiliary text. This might be the reason that ticket summaries mostly share similar structures to the descriptions so concatenating summaries might introduce duplicate information or even non-salient information. In addition, titles are generated at the creation of the ticket, so performing this would not delay the timing of when our algorithm can be applied in production. In general, the models fine-tuned with the prompt-prefix show strong performance thus demonstrate its effectiveness. Furthermore, we report the breakdown scores of the best model on D-Human dataset in Table \ref{tab:results_breakdown} in Appendix \ref{apdx:some_section}, which shows outstanding performance.

\subsection{Evaluation on Hard-to-Identify Tickets}
The performance on the D-Machine, D-Human, and D-Mixture is excellent, which demonstrates that the models learn well on these datasets. We further evaluate Ticket-BERT using human input on a set of hard-to-identify tickets that are difficult for annotators to identify in a short time. In particular, these tickets do not express specific incident issues thus are not easy to label by humans. We test fine-tuned Ticket-BERT model on dozens of these tickets. Surprisingly, we obtain nearly 90\% accuracy, which demonstrate the effectiveness of our proposed models in real industry scenarios. Moreover, we deploy the model on a real incident management system where our models are actively refine-tuned with newly collected data, newly developed ticket labels, and learned to deal with a diverse set of incoming tickets.

\begin{table}[]\small
\centering
\begin{adjustbox}{width=1\columnwidth}
\begin{tabular}{cccccc}
\hline
Dataset                    & Models    & Precision      & Recall         & F1-Score & AUC      \\ \hline
\multirow{6}{*}{D-Human}   & NB-BoW    & 71.07          & 73.74          & 69.87   & 0.955       \\
                           & NB-TF-IDF  & 78.18          & 76.67          & 76.74 & 0.973         \\
                           & LS-Model  & 84.36          & 84.08          & 83.96 & **          \\
                           & LR-BoW    & 85.16          & 84.12          & 84.58 & 0.98          \\
                           & LR-TF-IDF  & 85.81          & 82.89          & 83.98  & 0.986        \\
                           & Ticket-BERT & \textbf{86.40} & \textbf{85.50} & \textbf{85.90}& \textbf{0.988} \\ \hline
\multirow{6}{*}{D-Machine} & NB-BoW    & 89.17          & 91.98          & 89.71    & 0.997      \\
                           & NB-TF-IDF  & 92.28          & 93.72          & 92.69  & 0.999        \\
                           & LS-Model  & 96.01          & 94.52          & 95.14    & **      \\
                           & LR-BoW    & 97.91          & \textbf{98.07} & 97.98    & 1.000      \\
                           & LR-TF-IDF  & 97.15          & 97.50          & 97.30   & 1.000       \\
                           & Ticket-BERT & \textbf{98.42} & 97.98          & \textbf{98.19}& \textbf{1.000} \\ \hline
\multirow{6}{*}{D-Mixture} & NB-BoW    & 88.25          & 91.48          & 88.88   & 0.996       \\
                           & NB-TF-IDF  & 92.79          & 94.85          & 93.61  & 0.999        \\
                           & LS-Model  & 97.24          & 96.71          & 96.91 & **          \\
                           & LR-BoW    & 98.02          & \textbf{97.93} & 97.97 & 0.999          \\
                           & LR-TF-IDF  & 97.61          & 97.57          & 97.58  & 1.000        \\
                           & Ticket-BERT & \textbf{98.62} & 97.91          & \textbf{98.24}& \textbf{1.000} \\ \hline
\end{tabular}
\end{adjustbox}
\vspace{-.1in}
\caption{The performance on proposed datasets for baselines and Ticket-BERT models. **Please note that Language Studio does not provide per class probabilities, and as a result, we cannot calculate the AUC for this line.}
\vspace{-.1in}
\label{tab:results}
\end{table}

\section{Conclusion}
\label{sec:conclusion}
Ticket labeling is an essential task that provides an important summary of incident issues in support tickets and is critical to efficient triage to responsible teams for timely resolution. Existing methods are not applicable to this problem as they cannot sufficiently address the unique challenges in modern incident management systems (e.g., tickets are frequently manipulated or updated by machines or engineers who possess intrinsically different semantic structures). To address these shortcomings, we create three ticket datasets (i.e., D-Human, D-Machine, D-Mixture) to train domain-adaptive Ticket-BERT that achieves state-of-the-art predictions on all three. Our validation on a set of hard-to-identify tickets supplementally demonstrates Ticket-BERT’s effectiveness. While the model works well for IcM tickets for Azure Hardware, it would not generalize to all tickets within the Microsoft IcM system. However, we would be interested in collaborating and supporting additional efforts around generalizing this method for all IcM tickets. In the future, we will refine Ticket-BERT through an active learning cycle and make it adaptive to newly collected ticket data.

\begin{table}[!t]\small
% \begin{table}[!htbp]\small
\centering
\begin{adjustbox}{width=1\columnwidth}
\begin{tabular}{cccccc}
\hline
Dataset                                                                                          & Models    & Precision      & Recall         & F1-Score    & AUC   \\ \hline
\multirow{6}{*}{\begin{tabular}[c]{@{}l@{}}D-Human\\ +Title\end{tabular}}                  & NB-BoW    & 91.14          & 94.13          & 92.09     & 0.993     \\
  & NB-TF-IDF  & 93.46          & 94.23          & 93.77 & 0.997          \\
  & LS-Model  & 95.56          & 97.76          & 96.49  & **         \\
  & LR-BoW    & 97.92          & 97.90          & 97.91  & 0.999        \\
  & LR-TF-IDF  & 97.80          & 97.15          & 97.46 & 0.999          \\
  & Ticket-BERT & \textbf{98.76} & \textbf{99.17} & \textbf{98.96}& \textbf{1.000} \\ \hline
\multirow{6}{*}{\begin{tabular}[c]{@{}l@{}}D-Human\\+Title\\+Summary\end{tabular}} & NB-BoW    & 92.01          & 94.69          & 92.82 &   0.993      \\
  & NB-TF-IDF  & 94.30          & 95.03          & 94.61&    0.998      \\
  & LS-Model  & 97.81          & 97.75          & 97.77  &   **     \\
  & LR-BoW    & 98.24          & 98.15          & 98.19  &   0.998      \\
  & LR-TF-IDF  & 97.89          & 97.33          & 97.60 &  0.999       \\
  & Ticket-BERT & \textbf{98.36} & \textbf{98.88} & \textbf{98.61}& \textbf{1.000} \\ \hline
\multirow{6}{*}{\begin{tabular}[c]{@{}l@{}}D-Machine\\ +Title\end{tabular}}                  
    & NB-BoW    & 91.81          & 94.04          & 92.27   & 0.998       \\
    & NB-TF-IDF  & 94.73          & 96.10          & 95.16  & 0.999        \\
    & LS-Model  & 98.98          & 99.33          & 99.13   &**       \\
    & LR-BoW    & 98.96          & 99.23          & 99.09   & 1.000      \\
    & LR-TF-IDF  & 98.43          & 98.87          & 98.64  & 1.000       \\
    & Ticket-BERT & \textbf{99.31} & \textbf{99.24} & \textbf{99.27}& \textbf{1.000} \\ \hline
\multirow{6}{*}{\begin{tabular}[c]{@{}l@{}}D-Machine\\ +Title\\ +Summary\end{tabular}} 
    & NB-BoW    & 93.16          & 94.86          & 93.47  &  0.999    \\
    & NB-TF-IDF  & 95.08          & 96.84          & 95.70 &  1.000         \\
    & LS-Model  & 98.97          & 99.29          & 99.11  &  **      \\
    & LR-BoW    & 99.20          & 99.20          & 99.20  &  1.000        \\
    & LR-TF-IDF  & 98.92          & 99.02          & 98.97 &  1.000        \\
    & Ticket-BERT & \textbf{99.34} & \textbf{99.35} & \textbf{99.35}& \textbf{1.000} \\ \hline                                             
\multirow{6}{*}{\begin{tabular}[c]{@{}l@{}}D-Mixture\\ +Title\end{tabular}}                  
    & NB-BoW    & 91.87          & 94.56          & 92.60    & 0.998      \\
    & NB-TF-IDF  & 95.76          & 97.30          & 96.43   & 0.999      \\
    & LS-Model  & 98.07          & 97.82          & 97.91    & **     \\
    & LR-BoW    & 98.93          & 98.98          & 98.95    & 1.000     \\
    & LR-TF-IDF  & 98.67          & 98.78          & 98.72   & 1.000      \\
    & Ticket-BERT & \textbf{99.31} & \textbf{99.38} & \textbf{99.34}& \textbf{1.000} \\
                                                        \hline                                           
\multirow{6}{*}{\begin{tabular}[c]{@{}l@{}}D-Mixture\\  +Title\\ +Summary\end{tabular}}
    & NB-BoW    & 93.29          & 95.52          & 93.98   &  0.998     \\
    & NB-TF-IDF  & 95.51          & 97.45          & 96.34  &  0.999      \\
    & LS-Model  & 98.24          & 97.03          & 98.06   &   **    \\
    & LR-BoW    & 99.09          & 98.94          & 99.01   &  1.000     \\
    & LR-TF-IDF  & 98.75          & 98.67          & 98.71  &  1.000      \\
    & Ticket-BERT & \textbf{99.37} & \textbf{99.34} & \textbf{99.36} & \textbf{1.000} \\ \hline                                                                                                   
\end{tabular}
\end{adjustbox}
% \vspace{-.1in}
\caption{The performance on proposed datasets for baselines and Ticket-BERT models with auxiliary prompt-prefix.**Please note that Language Studio does not provide per class probabilities, and as a result, we cannot calculate the AUC for this line.}
\label{tab:results_prefix}
\end{table}

\bibliography{anthology,custom}

\begin{thebibliography}{24}
\expandafter\ifx\csname natexlab\endcsname\relax\def\natexlab#1{#1}\fi

\bibitem[{Ali~Zaidi et~al.(2022)Ali~Zaidi, Fraz, Shahzad, and
  Khan}]{ali2022multiapproach}
Syed~S Ali~Zaidi, Muhammad~Moazam Fraz, Muhammad Shahzad, and Sharifullah Khan.
  2022.
\newblock A multiapproach generalized framework for automated solution
  suggestion of support tickets.
\newblock \emph{International Journal of Intelligent Systems},
  37(6):3654--3681.

\bibitem[{Braşoveanu and Andonie(2020)}]{9373074}
Adrian M.~P. Braşoveanu and Răzvan Andonie. 2020.
\newblock \href {https://doi.org/10.1109/IV51561.2020.00051} {Visualizing
  transformers for nlp: A brief survey}.
\newblock In \emph{2020 24th International Conference Information Visualisation
  (IV)}, pages 270--279.

\bibitem[{Devlin et~al.(2018)Devlin, Chang, Lee, and
  Toutanova}]{devlin2018bert}
Jacob Devlin, Ming-Wei Chang, Kenton Lee, and Kristina Toutanova. 2018.
\newblock Bert: Pre-training of deep bidirectional transformers for language
  understanding.
\newblock \emph{arXiv preprint arXiv:1810.04805}.

\bibitem[{Gu et~al.(2011)Gu, Zhao, and Shu}]{gu2011analysis}
Hongying Gu, Long Zhao, and Chang Shu. 2011.
\newblock Analysis of duplicate issue reports for issue tracking system.
\newblock In \emph{The 3rd International Conference on Data Mining and
  Intelligent Information Technology Applications}, pages 86--91. IEEE.

\bibitem[{Gupta et~al.(2009)Gupta, Prasad, Luan, Rosu, and
  Ward}]{gupta2009multi}
Rajeev Gupta, K~Hima Prasad, Laura Luan, Daniela Rosu, and Chris Ward. 2009.
\newblock Multi-dimensional knowledge integration for efficient incident
  management in a services cloud.
\newblock In \emph{2009 IEEE International Conference on Services Computing},
  pages 57--64. IEEE.

\bibitem[{Gupta et~al.(2008)Gupta, Prasad, and Mohania}]{gupta2008automating}
Rajeev Gupta, K~Hima Prasad, and Mukesh Mohania. 2008.
\newblock Automating itsm incident management process.
\newblock In \emph{2008 International Conference on Autonomic Computing}, pages
  141--150. IEEE.

\bibitem[{Hadi et~al.(2018)Hadi, Al-Radaideh, and
  Alhawari}]{hadi2018integrating}
Wa'el Hadi, Qasem~A Al-Radaideh, and Samer Alhawari. 2018.
\newblock Integrating associative rule-based classification with na{\"\i}ve
  bayes for text classification.
\newblock \emph{Applied Soft Computing}, 69:344--356.

\bibitem[{Han and Akbari(2018)}]{han2018vertical}
Jianglei Han and Mohammad Akbari. 2018.
\newblock Vertical domain text classification: towards understanding it tickets
  using deep neural networks.
\newblock In \emph{Proceedings of the AAAI Conference on Artificial
  Intelligence}, volume~32.

\bibitem[{Herzig et~al.(2013)Herzig, Just, and Zeller}]{herzig2013s}
Kim Herzig, Sascha Just, and Andreas Zeller. 2013.
\newblock It's not a bug, it's a feature: how misclassification impacts bug
  prediction.
\newblock In \emph{2013 35th international conference on software engineering
  (ICSE)}, pages 392--401. IEEE.

\bibitem[{Kadhim(2019)}]{kadhim2019survey}
Ammar~Ismael Kadhim. 2019.
\newblock Survey on supervised machine learning techniques for automatic text
  classification.
\newblock \emph{Artificial Intelligence Review}, 52(1):273--292.

\bibitem[{Kallis et~al.(2019)Kallis, Di~Sorbo, Canfora, and
  Panichella}]{kallis2019ticket}
Rafael Kallis, Andrea Di~Sorbo, Gerardo Canfora, and Sebastiano Panichella.
  2019.
\newblock Ticket tagger: Machine learning driven issue classification.
\newblock In \emph{2019 IEEE International Conference on Software Maintenance
  and Evolution (ICSME)}, pages 406--409. IEEE.

\bibitem[{Khan et~al.(2021)Khan, Naseer, Hayat, Zamir, Khan, and
  Shah}]{khan2021transformers}
Salman Khan, Muzammal Naseer, Munawar Hayat, Syed~Waqas Zamir, Fahad~Shahbaz
  Khan, and Mubarak Shah. 2021.
\newblock Transformers in vision: A survey.
\newblock \emph{ACM Computing Surveys (CSUR)}.

\bibitem[{Khowongprasoed and Titijaroonroj(2022)}]{9836250}
Kraidet Khowongprasoed and Taravichet Titijaroonroj. 2022.
\newblock \href {https://doi.org/10.1109/JCSSE54890.2022.9836250} {Automatic
  thai ticket classification by using machine learning for it infrastructure
  company}.
\newblock In \emph{2022 19th International Joint Conference on Computer Science
  and Software Engineering (JCSSE)}, pages 1--6.

\bibitem[{Kim and Shawe-Taylor(1992)}]{kim1992fast}
Jong~Yong Kim and John Shawe-Taylor. 1992.
\newblock Fast multiple keyword searching.
\newblock In \emph{Annual Symposium on Combinatorial Pattern Matching}, pages
  41--51. Springer.

\bibitem[{Kowsari et~al.(2019)Kowsari, Jafari~Meimandi, Heidarysafa, Mendu,
  Barnes, and Brown}]{kowsari2019text}
Kamran Kowsari, Kiana Jafari~Meimandi, Mojtaba Heidarysafa, Sanjana Mendu,
  Laura Barnes, and Donald Brown. 2019.
\newblock Text classification algorithms: A survey.
\newblock \emph{Information}, 10(4):150.

\bibitem[{Lewis(1995)}]{lewis1995sequential}
David~D Lewis. 1995.
\newblock A sequential algorithm for training text classifiers: Corrigendum and
  additional data.
\newblock In \emph{Acm Sigir Forum}, volume~29, pages 13--19. ACM New York, NY,
  USA.

\bibitem[{Maksai et~al.(2014)Maksai, Bogojeska, and
  Wiesmann}]{maksai2014hierarchical}
Andrii Maksai, Jasmina Bogojeska, and Dorothea Wiesmann. 2014.
\newblock Hierarchical incident ticket classification with minimal supervision.
\newblock In \emph{2014 IEEE International Conference on Data Mining}, pages
  923--928. IEEE.

\bibitem[{Mukunthan and Selvakumar(2019)}]{mukunthan2019multilevel}
MA~Mukunthan and S~Selvakumar. 2019.
\newblock Multilevel petri net-based ticket assignment and it management for
  improved it organization support.
\newblock \emph{Concurrency and Computation: Practice and Experience},
  31(14):e5297.

\bibitem[{Revina et~al.(2020)Revina, Buza, and Meister}]{revina2020ticket}
Aleksandra Revina, Krisztian Buza, and Vera~G Meister. 2020.
\newblock It ticket classification: the simpler, the better.
\newblock \emph{IEEE Access}, 8:193380--193395.

\bibitem[{Settles(2009)}]{settles2009active}
Burr Settles. 2009.
\newblock Active learning literature survey.

\bibitem[{Sidhu and Prasanna(2001)}]{sidhu2001fast}
Reetinder Sidhu and Viktor~K Prasanna. 2001.
\newblock Fast regular expression matching using fpgas.
\newblock In \emph{The 9th Annual IEEE Symposium on Field-Programmable Custom
  Computing Machines (FCCM'01)}, pages 227--238. IEEE.

\bibitem[{Son et~al.(2014)Son, Hazlewood, and Peterson}]{son2014automating}
Gwang Son, Victor Hazlewood, and Gregory~D Peterson. 2014.
\newblock On automating xsede user ticket classification.
\newblock In \emph{Proceedings of the 2014 Annual Conference on Extreme Science
  and Engineering Discovery Environment}, pages 1--7.

\bibitem[{Yao et~al.(2019)Yao, Mao, and Luo}]{yao2019clinical}
Liang Yao, Chengsheng Mao, and Yuan Luo. 2019.
\newblock Clinical text classification with rule-based features and
  knowledge-guided convolutional neural networks.
\newblock \emph{BMC medical informatics and decision making}, 19(3):31--39.

\bibitem[{Zhou et~al.(2016)Zhou, Tong, Gu, and Gall}]{zhou2016combining}
Yu~Zhou, Yanxiang Tong, Ruihang Gu, and Harald Gall. 2016.
\newblock Combining text mining and data mining for bug report classification.
\newblock \emph{Journal of Software: Evolution and Process}, 28(3):150--176.

\end{thebibliography}

\newpage
\appendix
\section{Appendices}
\label{sec:appendix}

\subsection{Per-class performance for Ticket-BERT}

\label{apdx:appendix_subfig_confusion_matrix}
\begin{figure}[H]
	\centering
	\begin{subfigure}%
		\centering
		\includegraphics[width=\linewidth]{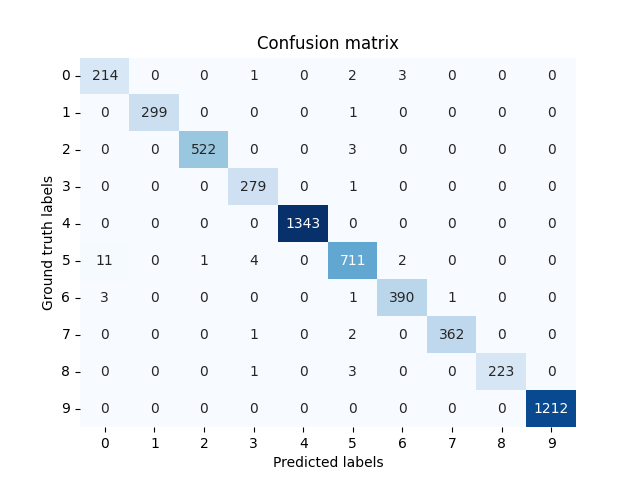}%
		\caption{Ticket-BERT with D-human}
	\end{subfigure}%

	%\vspace*{2pt}%
	
	\begin{subfigure}%
		\centering
		\includegraphics[width=\linewidth]{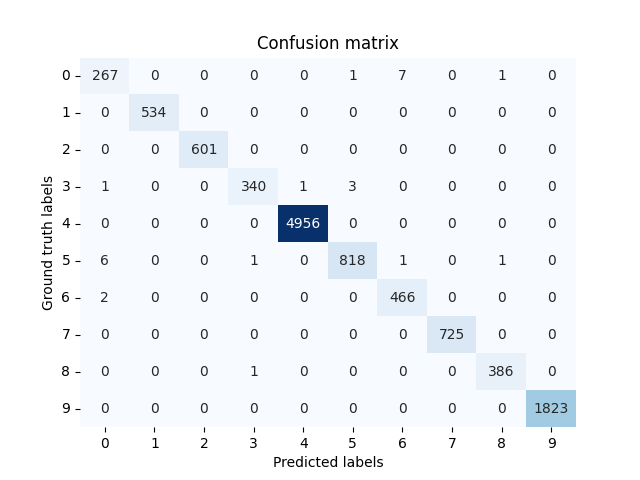}%
		\caption{Ticket-BERT with D-machine}
	\end{subfigure}%

	%\vspace*{2pt}%
	
	\begin{subfigure}%
		\centering
		\includegraphics[width=\linewidth]{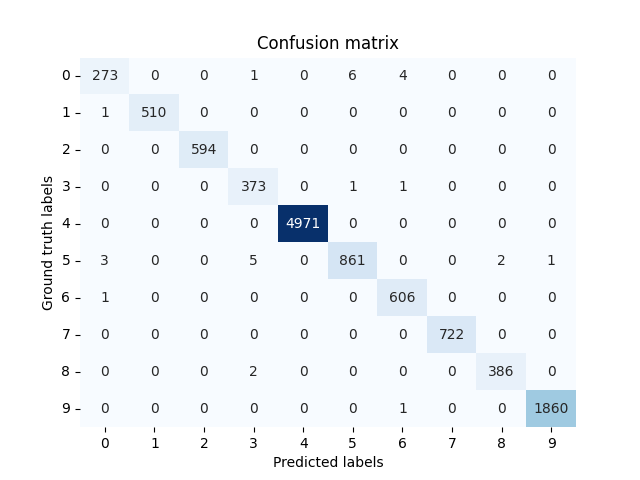}%
		\caption{Ticket-BERT with D-mixture}
	\end{subfigure}%

	%\caption{\label{apdx:appendix_subfig1}A pair of images}
\end{figure}%

\subsection{ROC-AUC and auPRC curves for Ticket-BERT}

\label{apdx:appendix_subfig_roc_auc}

\begin{figure}[H]
	\centering
	\begin{subfigure}%
		\centering
		\includegraphics[width=1.05\linewidth]{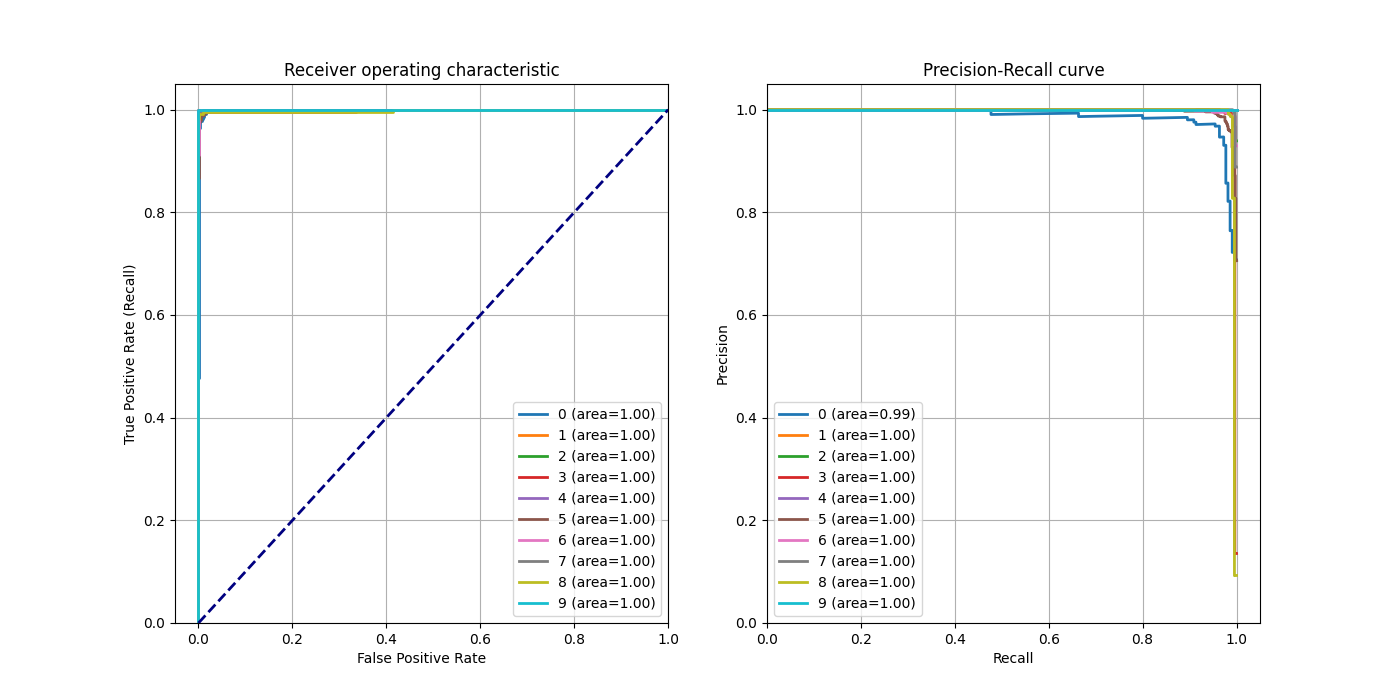}%
		\caption{Ticket-BERT with D-human}
	\end{subfigure}%

	\vspace*{4pt}%
	
	\begin{subfigure}%
		\centering
		\includegraphics[width=1.05\linewidth]{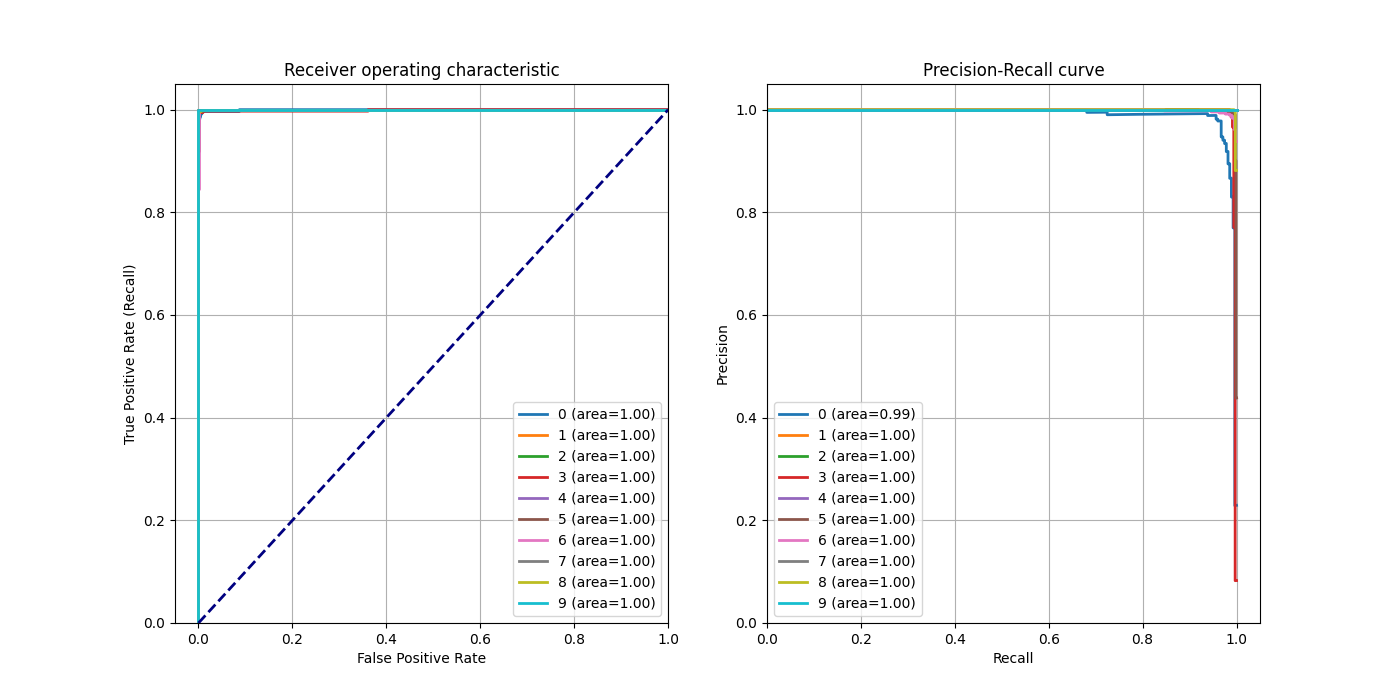}%
		\caption{Ticket-BERT with D-machine}
	\end{subfigure}%

	\vspace*{4pt}%
	
	\begin{subfigure}%
		\centering
		\includegraphics[width=1.05\linewidth]{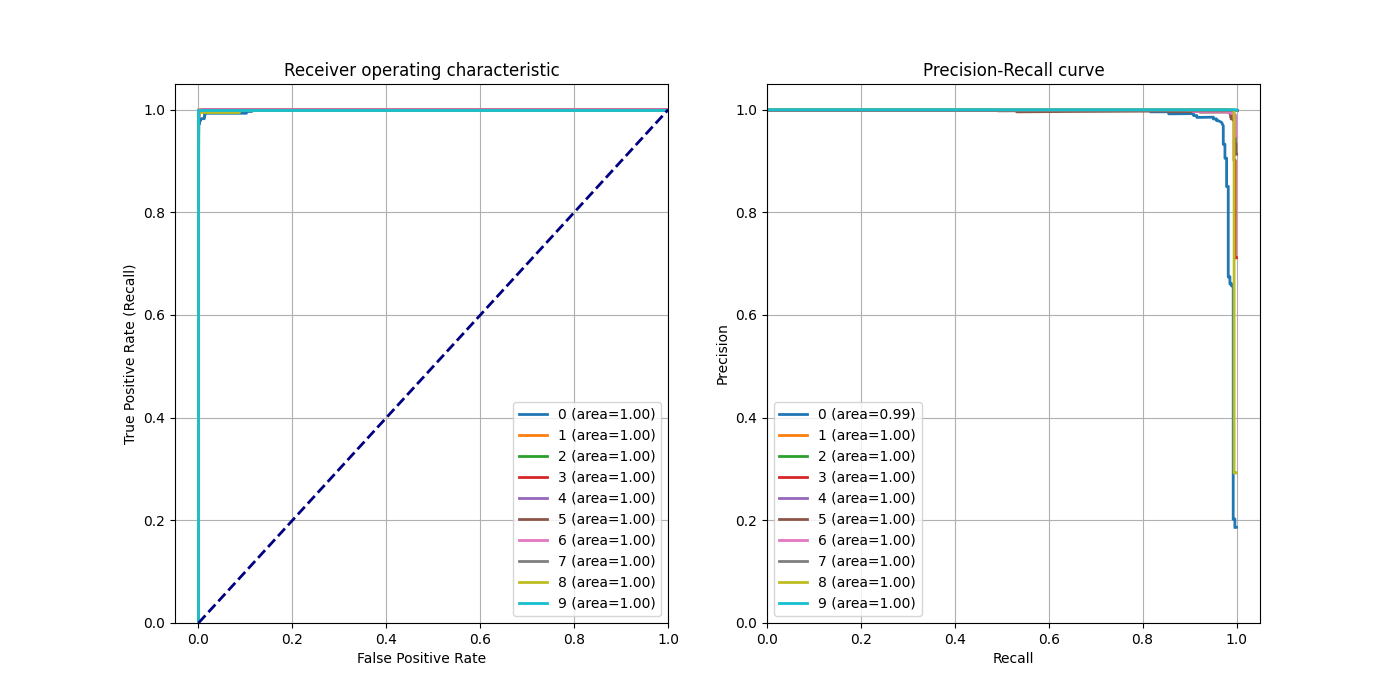}%
		\caption{Ticket-BERT with D-mixture}
	\end{subfigure}%

    \vspace*{10pt}%
    
    \begin{tabular}{cc}
\hline
\textbf{Label Indicator}                    & \textbf{Label Name}  \\ \hline
        0    & Buildout \\ \hline
        1    & GDCO Escalation \\ \hline
        2    & SKU Artifacts \\ \hline
        3    & Customer Facing RCA \\ \hline
        4    & CloudNet FPGA Alerts \\ \hline  
        5    & Capacity Recovery Requests \\ \hline  
        6    & ECO Request \\ \hline  
        7    & Anvil Rate \\ \hline  
        8    & Firmware Deployment \\ \hline 
        9    & Attestation \\ \hline
\end{tabular}

	%Class labels are: 0 - Buildout, 1. GDCOApp Ticket Escalation, 2. SKU Artifacts, 3. Customer Facing RCA Requests, 4. CloudNet FPGA Alerts, 5. Capacity Recovery Requests, 6. ECO Request or HW Profile Support, 7. Anvil Rate Limit Hit, 8. Fireware Deployment, 9. Attestation
\end{figure}%

\vspace*{10pt}%
\subsection{Ticket Examples and Breakdown Results for Ticket Labeling}

\label{apdx:some_section}

\begin{table*}[]
\centering
\def\tabularxcolumn#1{m{#1}}
\begin{adjustbox}{width=0.7\textwidth}
\begin{tabularx}{0.8\textwidth}{c|X}
\hline
Time                              & Ticket Descriptions                                         \\\hline
2020-05-26 & Acknowledging incident
 \\\hline
2020-06-03 & Hi Ben, I've just taken this IcM (not as a dev) but to determine next steps/team. This seems to have a lot of churn so we are hoping to reduce it...                   \\\hline
2020-06-15 & If the goal is to move the from AU (legacy boot mode) to AO flavor (UEFI boot mode), there should be an issue... \\\hline
2020-07-25 & Vijay owns remediation updates during OOS; please take a look. This ICM may be too old - please resolve if things are OK now. Thanks!
                        \\\hline
\end{tabularx}
\end{adjustbox}
% \vspace{-.1in}
\caption{An example of parts of ticket description updates.}
\vspace{-.1in}
\label{tab:ticket_timeline}
\end{table*}

\label{apdx:some_example}
\begin{table*}[]
\centering
\def\tabularxcolumn#1{m{#1}}
\begin{adjustbox}{width=0.65\textwidth}
\begin{tabularx}{0.8\textwidth}{c|X|X}
\hline
 Update & Human entered/updated ticket descriptions  & Machine generated/updated ticket descriptions \\ \hline
 $T_1$ & @NAMEMASKED - Assigning to you to begin investigation into this   issue.                           & {[}automated{]} Update: Node is currently out-for-repair. Need to validate   proper repair action. \\ \hline
  $T_2$ & If the goal is to move the from   AU (legacy boot mode) to AO flavor (UEFI boot mode), there should be …  & {[}automated{]} Update: This IcM has not been updated in the past 7 days.                          \\ \hline
  $T_3$ & All the Nodes are in Production state after Bios update to the right   supported Bios version \& flavor. Hence closing the ICM. & The severity in this IcM incident is inherited from the severity of the   GDCO Ticket              \\ \hline
\end{tabularx}
\end{adjustbox}
% \vspace{-.1in}
\caption{Human-entered ticket descriptions (left) v.s. machine-generated ticket descriptions (right).}
\label{tab:ticket_examples}
\end{table*}

\begin{table*}[]
\centering
\begin{adjustbox}{width=1.3\columnwidth}
\begin{tabular}{lcccccc}
\hline
Labels          & Precision & Recall & F1   Score &AUC\\ \hline
Buildout             & 93.18     & 93.18  & 93.18  & 1.000        \\
GDCOApp Ticket Escalation              & 99.67     & 99.33  & 99.50     & 1.000     \\
SKU Artifacts              & 99.24     & 99.43  & 99.33   & 1.000       \\
Customer Facing RCA Requests              & 96.40     & 95.71  & 96.06  & 1.000       \\
CloudNet FPGA Alerts              & 99.48     & \textbf{99.93}  & 99.70    & 1.000      \\
Capacity Recovery Requests              & 98.05     & 96.57  & 97.30      & 1.000     \\
ECO Request or HW Profile Support              & 96.55     & 99.24  & 97.88    & 1.000       \\
Anvil Rate Limit Hit              & \textbf{100.00}    & 99.73  & 99.86    & 1.000     \\
Firmware Deployment              & 99.11     & 98.24  & 98.67    & 1.000      \\
Attestation              & \textbf{100.00}    & 99.92  & \textbf{99.96}   & 1.000      \\ \hline
\end{tabular}
\end{adjustbox}
\caption{The breakdown performance on D-Human dataset for the best TicketBERT model. See Appendix \ref{apdx:appendix_subfig_confusion_matrix} \ref{apdx:appendix_subfig_roc_auc} for ROC curve and detailed confusion matrix for the best D-Human, D-Machine, D-Mixture Ticket-BERT models.}
\label{tab:results_breakdown}
\end{table*}

\newpage

\end{document}